\documentclass[lettersize,journal]{IEEEtran}
\usepackage{amsmath,amsfonts}
\usepackage{algorithmic}
\usepackage{array}
\usepackage[caption=false,font=normalsize,labelfont=sf,textfont=sf]{subfig}
\usepackage{textcomp}
\usepackage{stfloats}
\usepackage{url}
\usepackage{verbatim}
\usepackage{graphicx}

\usepackage{dsfont}
\usepackage{booktabs, multirow}
\usepackage{marvosym, hyperref}
\usepackage{cleveref}
\usepackage{xcolor}

\def\BibTeX{{\rm B\kern-.05em{\sc i\kern-.025em b}\kern-.08em
    T\kern-.1667em\lower.7ex\hbox{E}\kern-.125emX}}
\usepackage{balance}
\begin{document}
\title{Compact 3D Gaussian Splatting for Static and Dynamic Radiance Fields}
\author{Joo Chan Lee, \textit{Graduate Student Member, IEEE}, Daniel Rho, Xiangyu Sun, Jong Hwan Ko, \textit{Member, IEEE}, and Eunbyung Park, \textit{Member, IEEE}
\thanks{Manuscript received August, 2024; This work was supported in part by the Institute of Information and Communications Technology Planning and Evaluation (IITP) funded by the Korea government (MSIT) under Grant RS-2021-II212068 (Artificial Intelligence Innovation Hub), Grant RS-2019-II190421 (AI Graduate School Support Program (Sungkyunkwan University)), and Grant No. 2018-0-00207, RS-2018-II180207 (Immersive Media Research Laboratory); and in part by the Culture, Sports, and Tourism R\&D Program through the Korea Creative Content Agency funded by the Ministry of Culture, Sports and Tourism in 2024 under Grant RS-2024-00348469 (Research on neural watermark technology for copyright protection of generative AI 3D content). \textit{(Corresponding author: Eunbyung Park; Jong Hwan Ko.)}

Joo Chan Lee is with the Department of Artificial Intelligence, Sungkyunkwan University, Suwon 16419, South Korea (e-mail: maincold2@skku.edu).

Daniel Rho is with the Department of Computer Science, University of North Carolina at Chapel Hill, NC 27599, USA and KT, Seoul 06763, South Korea (e-mail: dnl03c1@cs.unc.edu).

Xiangyu Sun is with the Department of Electrical and Computer Engineering, Sungkyunkwan University, Suwon 16419, South Korea (e-mail: xiangyusun@g.skku.edu).

Jong Hwan Ko and Eunbyung Park are with the Department of Electronic and Electrical Engineering, Sungkyunkwan University, Suwon 16419, South Korea (e-mail: jhko@skku.edu; epark@skku.edu).
}}

\markboth{Preprint}{}
% \markboth{Journal of \LaTeX\ Class Files,~Vol.~18, No.~9, September~2020}%
% {How to Use the IEEEtran \LaTeX \ Templates}

\maketitle

\begin{abstract}
Neural Radiance Fields (NeRFs) have demonstrated remarkable potential in capturing complex 3D scenes with high fidelity. However, one persistent challenge that hinders the widespread adoption of NeRFs is the computational bottleneck due to the ray-wise volumetric rendering. On the other hand, 3D Gaussian splatting (3DGS) has recently emerged as an alternative representation that leverages a 3D Gaussian-based representation and introduces an approximated volumetric rendering, achieving very fast rendering speed and promising image quality. Furthermore, subsequent studies have successfully extended 3DGS to dynamic 3D scenes, demonstrating its wide range of applications. However, a significant drawback arises as 3DGS and its following methods entail a substantial number of Gaussians to maintain the high fidelity of the rendered images, which requires a large amount of memory and storage. To address this critical issue, we place a specific emphasis on two key objectives: reducing the number of Gaussian points without sacrificing performance and compressing the Gaussian attributes, such as view-dependent color and covariance. To this end, we propose a learnable mask strategy that significantly reduces the number of Gaussians while preserving high performance. In addition, we propose a compact but effective representation of view-dependent color by employing a grid-based neural field rather than relying on spherical harmonics. Finally, we learn codebooks to compactly represent the geometric and temporal attributes by residual vector quantization. With model compression techniques such as quantization and entropy coding, we consistently show over 25$\times$ reduced storage and enhanced rendering speed compared to 3DGS for static scenes, while maintaining the quality of the scene representation. For dynamic scenes, our approach achieves more than 12$\times$ storage efficiency and retains a high-quality reconstruction compared to the existing state-of-the-art methods. Our work provides a comprehensive framework for both static and dynamic 3D scene representation, achieving high performance, fast training, compactness, and real-time rendering. Our project page is available at \href{https://maincold2.github.io/c3dgs/}{https://maincold2.github.io/c3dgs/}.
\end{abstract}


\section{Introduction}

\begin{figure*}[t]
    \begin{center}
    \includegraphics[width=1.0\textwidth]{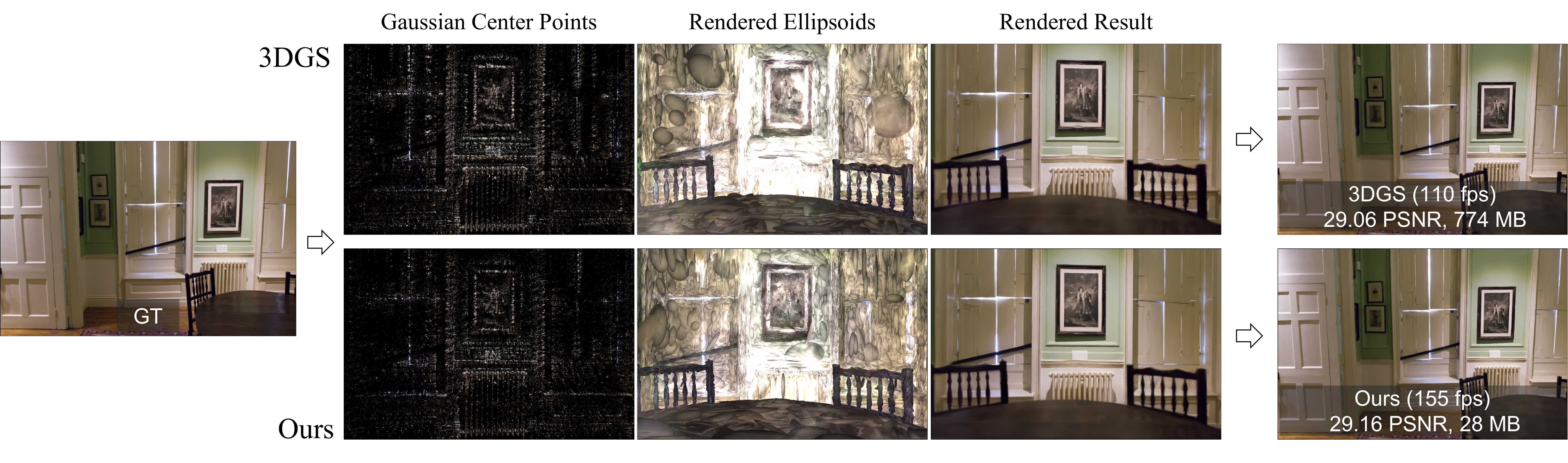}
    \end{center}
    % \vspace{-2em}
    \caption{Our method achieves reduced storage and faster rendering speed while maintaining high-quality renderings of 3DGS~\cite{3dgs}. The core idea is to effectively remove the redundant Gaussians that do not significantly contribute to the overall performance (the sparser distribution of Gaussian points and reduced ellipsoid redundancy shown in the figure). We also introduce a more compact representation of Gaussian attributes, resulting in markedly improved storage efficiency and rendering speed.}
    % \vspace{-1.5em}
\label{fig:demo}
\end{figure*}

\IEEEPARstart{T}{he} field of neural rendering has witnessed substantial advancements in recent years, driven by the pursuit of rendering photorealistic 3D scenes from limited input data. Among the pioneering approaches, Neural Radiance Field (NeRF)~\cite{nerf} has gained considerable attention for its remarkable ability to generate high-fidelity images and 3D reconstructions of scenes from only a collection of 2D images. Follow-up research efforts have been dedicated to improving image quality~\cite{mip-nerf, instant-ngp}, accelerating training and rendering speed~\cite{killonerf, fridovich2022plenoxels, tensorf, kplanes, instant-ngp, dsnerf}, reducing memory and storage footprints~\cite{masked, VQAD, cam}, and expanding its use to dynamic 3D scenes~\cite{dnerf, Li_2021_CVPR, Li_2022_CVPR, kplanes, mixvoxels}.

Despite the massive efforts, one persistent challenge that hinders the widespread adoption of NeRFs is the computational bottleneck due to pixel-wise volumetric rendering. 
Since it demands dense point sampling along the ray to render a pixel, which requires significant computational resources, NeRFs often fail to achieve real-time rendering on hand-held devices or low-end GPUs.
This challenge limits their use in practical scenarios where fast rendering speed is essential, such as various interactive 3D applications.

3D Gaussian splatting (3DGS)~\cite{3dgs} has emerged as an alternative representation that can achieve both real-time rendering and high rendering quality.
This approach leverages a point-based representation associated with 3D Gaussian attributes and adopts the rasterization pipeline to render the images. % rather than volumetric rendering.
Highly optimized customized cuda kernels to maximize the parallelism and clever algorithmic tricks enable unprecedented rendering speed without compromising the image quality.
Not confined to static 3D scenes, several works~\cite{4dgs_kplanes, realtime4dgs, stg} have demonstrated the applicability of 3DGS also in dynamic 3D scenes.
Despite increased temporal dimension, they also achieved fast rendering with high representation quality as well as in static 3D scenes.

However, these 3D Gaussian-based rendering methods share a significant drawback: they require a large amount of memory and storage (e.g., 3DGS often needs over 1GB to represent a static real-world scene).
This is primarily due to the necessity of a substantial number of Gaussians to ensure high-quality images (Fig.~\ref{fig:demo}).
Moreover, each Gaussian has many attributes, such as position, scale, rotation, color, and opacity, requiring numerous parameters.
For dynamic scenes, the requirement to model temporal movements introduces extra attributes, further increasing the storage overheads.

In this work, we propose a compact 3D Gaussian representation framework that can enhance memory and storage efficiency while attaining high reconstruction quality, fast training speed, and real-time rendering (Fig.~\ref{fig:demo}).
Our proposed method is an end-to-end framework, broadly applicable to 3DGS-based methods and primarily improving efficiency in two key areas: the number of Gaussians and the average size of each Gaussian.
First, we reduce the total number of Gaussians using learnable masks without sacrificing representation performance.
The densification process of 3DGS or 3DGS-based methods, consisting of cloning and splitting Gaussians, increases the number of Gaussians, and this is a crucial component in achieving a high level of detail.
However, we observed that the current densification algorithm produces myriads of redundant and insignificant Gaussians, resulting in high memory and storage requirements.
We propose a novel learnable masking strategy for Gaussian Splatting in both static and dynamic scenes, designed to identify and eliminate non-essential Gaussians that contribute minimally to the overall rendering quality.
With the proposed masking method, we can reduce the number of Gaussians based on their volume and transparency during training, while achieving high performance.
In addition to the efficient memory and storage usage, we can achieve faster rendering speed since the computational complexity of the rendering process in 3DGS is highly correlated to the number of Gaussians.

Second, we decrease the average size of each Gaussian by compressing the Gaussian attributes, such as view-dependent color, covariance, and temporal attributes. 
In 3DGS, each Gaussian has its own attributes, and it does not exploit spatial redundancy, which has been widely utilized for various types of signal compression.
For example, neighboring Gaussians may share similar color attributes, and we can reuse similar colors from neighboring Gaussians.
Given this motivation, we incorporate a grid-based neural field to efficiently represent view-dependent colors rather than using per-Gaussian color attributes.
When provided with the query Gaussian points, we extract the color attribute from the compact grid representation, avoiding the need to store it for each Gaussian separately.
For our initial approach, we opt for a hash-based grid representation (Instant NGP~\cite{instant-ngp}) from among several candidates due to its compactness and fast processing speed. This choice has led to a significant reduction in the spatial complexity of the previous methods.

In addition, 3DGS-based approaches represent a scene with numerous small Gaussians collectively, and each Gaussian primitive is not expected to show high diversity. 
We found that the majority of Gaussians exhibit similar geometry in both static and dynamic scenes, with limited variation in scale and rotation.
Based on this observation, we propose a codebook-based approach for modeling the geometry of Gaussians.
It learns to find similar patterns or geometry shared across each scene and only stores the codebook index for each Gaussian, resulting in a very compact representation.
Moreover, we found that a compact codebook suffices to represent a highly detailed scene, therefore, the spatial and computational overhead can be insignificant.
Similarly, in a dynamic scene, Gaussians may exhibit similar motion trajectories. For instance, groups of moving parts typically follow similar motion patterns, while static areas remain motionless.
Therefore, we also learn codebooks to represent temporal attributes of dominant motions.

We have extensively tested our proposed compact Gaussian representation on various datasets.
In end-to-end training, our approach consistently showed reduced storage (15 $\times$ less than 3DGS for static scenes and 9 $\times$ less than STG~\cite{stg} for dynamic scenes) and enhanced rendering speed, all while maintaining the quality of the scene representation.
Furthermore, our method can benefit from simple post-processing techniques such as quantization and entropy coding, consequently achieving over 25$\times$ and 12$\times$ compression for static and dynamic scenes, respectively.

The earlier version of this research~\cite{c3dgs} was published at CVPR 2024 as a highlight presentation, focusing on a compact Gaussian representation for static scenes. This updated version broadens the scope to include dynamic scenes with significant enhancements: 1) We successfully extend the learnable masking approach for Gaussians moving over time, demonstrating its wide applicability.
For static scenes, several methods~\cite{lightgaussian, morton3d} have attempted to estimate and remove non-essential Gaussians after training, yielding promising results.
However, removing non-essential Gaussians after training has been more challenging in dynamic scenes, as it requires measuring the importance of each Gaussian over the entire duration.
In contrast, our proposed masking strategy simplifies the process by eliminating such complexities, learning the actual rendering impact of each Gaussian across all timestamps during training iterations through gradient descent.
2) To compactly represent the motions of Gaussians, we propose learning representative temporal trajectories by applying the codebook-based approach to temporal attributes. We successfully represent temporal attributes parameter-efficiently and validate that other compact representations for geometry and color are applicable for dynamic scenes as well as for static scenes.
3) Extensive experiments and analysis demonstrate the effectiveness of our approach in dynamic settings. We achieve more than a tenfold increase in parametric efficiency compared to STG, the state-of-the-art method for dynamic scene representation, while maintaining comparable performance.

\section{Related Work}
\subsection{Neural Rendering for 3D Scenes}
\subsubsection{Neural Radiance Fields}
Neural radiance fields (NeRFs) have significantly expanded the horizons of 3D scene reconstruction and novel view synthesis. NeRF~\cite{nerf} introduced a novel approach to synthesizing novel views of 3D scenes, representing volume features by utilizing Multilayer Perceptrons (MLPs) and introducing volumetric rendering. Since its inception, various works have been proposed to enhance performance in diverse scenarios, such as different resolutions of reconstruction~\cite{mip-nerf, mip360}, the reduced number of training samples~\cite{yu2021pixelnerf, niemeyer2022regnerf, sinnerf, ibrnet, depthprior}, and reconstruction of large realistic scenes~\cite{rawnerf, blocknerf} and dynamic scenes~\cite{dnerf,Li_2021_CVPR, Li_2022_CVPR, Gao_2021_ICCV}. However, NeRF's reliance on MLP has been a bottleneck, particularly causing slow training and inference.

In an effort to address the limitations, grid-based methods emerged as a promising alternative. These approaches using explicit voxel grid structures~\cite{fridovich2022plenoxels, DVGO, zipnerf, tineuvox, nsvf, devrf} have demonstrated a significant improvement in training speed compared to traditional MLP-based NeRF methods. Nevertheless, despite this advancement, grid-based methods still suffer from relatively slow inference speeds and, more importantly, require large amounts of memory. This has been a substantial hurdle in advancing towards more practical and widely applicable solutions.

Subsequent research efforts have been directed toward the reduction of the memory footprint while maintaining or even enhancing the performance quality by grid factorization~\cite{tensorf, EG3D, kplanes, multensorf, strivec, mipgrid}, hash grids~\cite{instant-ngp, howfar}, grid quantization~\cite{VQAD, birf} or pruning~\cite{fridovich2022plenoxels, masked}. These methods have also been instrumental in the fast training of 3D scene representation, thereby making more efficient use of computational resources. However, a persistent challenge that remains is the ability to achieve real-time rendering of large-scale scenes. The volumetric sampling inherent in these methods, despite their advancements, still poses a limitation.

\subsubsection{Point-based Rendering and Radiance Field}
To achieve high computational efficiency, Point-NeRF~\cite{pointnerf} proposed rendering with discrete points rather than continuous fields.
NeRF-style volumetric rendering and point-based $\alpha$-blending fundamentally share the same model for rendering images but differ significantly in their rendering algorithms~\cite{3dgs}. 
NeRFs offer a continuous feature representation of the entire volume as empty or occupied spaces, which necessitate costly volumetric sampling to render a pixel, leading to high computational demands.
In contrast, points provide an unstructured, discrete representation of a volume geometry by the creation, destruction, and movement of points, and a pixel is rendered by blending several ordered points overlapping the pixel.
By optimizing opacity and positions~\cite{kopanas2021}, point-based approaches can achieve fast rendering while avoiding the drawbacks of sampling in a continuous space.

Point-based methods have been widely used in rendering 3D scenes, where the simplest form is point clouds. 
However, point clouds can lead to visual artifacts such as holes and aliasing.
To mitigate this, point-based neural rendering methods have been proposed, processing the points through rasterization-based point splatting and differentiable rasterization~\cite{yifan, wiles, pulsar}.
The points were represented by neural features and rendered with CNNs~\cite{aliev, kopanas2021, meshry}.
However, these methods heavily rely on Multi-View Stereo (MVS) for initial geometry, inheriting its limitations, especially in challenging scenarios like areas lacking features, shiny surfaces, or fine structures.

Neural Point Catacaustics~\cite{kopanas2022} addressed the issue of view-dependent effect through the use of an MLP, yet it still depends on MVS geometry for its input. 
Without the need for MVS, Zhang et al.\cite{zhang2022} incorporated Spherical Harmonics (SH) for directional control. However, this method is constrained to managing scenes with only a single object and requires the use of masks during its initialization phase. 
Recently, 3D Gaussian Splatting (3DGS)~\cite{3dgs} proposed using 3D Gaussians as primitives for real-time neural rendering, opening up a new paradigm for 3D scene rendering~\cite{survey1, survey2}.
3DGS utilizes highly optimized custom CUDA kernels and ingenious algorithmic approaches to achieve unparalleled rendering speed without sacrificing image quality.

\begin{figure*}[t]
    \begin{center}
    \includegraphics[width=1.0\linewidth]{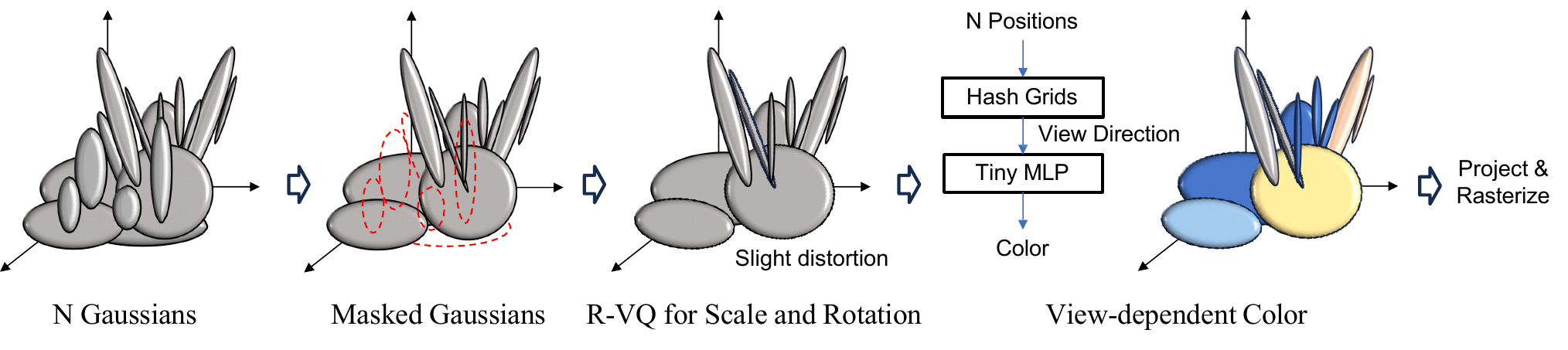}
    \end{center}
    % \vspace{-1.5em}
    \caption{The detailed architecture of our proposed compact 3D Gaussian.}
    % \vspace{-1.5em}
\label{fig:arch}
\end{figure*}

While 3DGS does not require dense sampling for each ray, it does require a substantial number of 3D Gaussians to maintain a high level of quality in the resulting rendered images.
Additionally, since each Gaussian consists of several rendering-related attributes like covariance matrices and SH with high degrees, 3DGS demands significant memory and storage resources, e.g., exceeding 1GB for a realistic scene.
Our work aims to alleviate this parameter-intensive requirement while preserving high rendering quality, fast training, and real-time rendering.

\subsubsection{Concurrent works}
Several recent works have pursued storage-efficient 3DGS, similar to our objective.
These approaches used conventional compression techniques such as scalar or vector quantization~\cite{lightgaussian, compact3d, morton3d, c3d2d, eagles} and entropy coding~\cite{morton3d, c3d2d}, while some of them involve pruning 3D Gaussians based on the significance assessed post-training~\cite{lightgaussian, morton3d}.
Despite the incorporation of a line of compression techniques, only EAGLES~\cite{eagles} and our method support end-to-end training.
Whereas EAGLES controlled the number of Gaussians by just adjusting the densification schedule, resulting in a sub-optimal reduction, our approach is the only work that successfully masks out ineffective Gaussians during training. 

\subsection{Neural Rendering for Dynamic Scenes}
Neural rendering literature has evolved to capture the temporal changes of dynamic scenes, building on the pioneering methods proposed for static environments.
\subsubsection{NeRF-based Methods}
Several extensions of NeRF initially attempted to represent sparser dynamic scenes from a monocular video~\cite{dnerf,Li_2021_CVPR, tineuvox, Gao_2021_ICCV, Du_2021_ICCV, dynibar, Liu_2023_CVPR, nerfies, hypernerf, Tretschk_2021_ICCV}, where a single camera captures the scene from one perspective per timestamp.
These methods generally utilize scene flow or depth information to overcome the limited supervision from single viewpoints.
Although these approaches have shown promising advancements in non-rigid scenarios, reconstructing large-scale dynamic scenes remains challenging when relying on monocular videos.

To reconstruct complex dynamic scenes in practical applications, recent works have utilized synchronized multi-view videos, which offer detailed supervision from various viewpoints and timestamps.
DyNeRF~\cite{Li_2022_CVPR} integrates NeRF with time-conditioned latent codes, successfully representing more complicated real-world scenes.
However, DyNeRF demands extensive training and rendering times, yielding sub-optimal performance.
Subsequent studies have explored resolving its inefficiency by focusing on motion between frames~\cite{streamrf}, efficient sampling~\cite{hyperreel}, or decomposing scene components~\cite{nerfplayer, mixvoxels, Wang_2023_CVPR}.
In addition, several methods have adopted efficient grid structures, which have succeeded for static scenes, such as factorized~\cite{hexplanes, kplanes, tensor4d, humanrf} or hash~\cite{maskedhash} grids.
However, reliance on the dense sampling of those NeRF-based methods still poses challenges in achieving real-time rendering for large real-world scenes.

\subsubsection{3DGS-based Methods}
More recently, several works have extended 3DGS to dynamic scenes.
Dynamic3D~\cite{dynamic3dgs} constructs a sequence of 3D Gaussians, representing the positions and rotations discretely at each timestamp.
To improve efficiency in modeling temporal movements, a line of works used additional architectures such as MLP~\cite{deformable4dgs} or grids~\cite{4dgs_kplanes, 4k4d}, following the NeRF paradigm.
Another line of research learned additional parameters as coefficients for various bases, such as linear~\cite{realtime4dgs}, polynomial~\cite{gaussianflow, stg}, Fourier~\cite{fourier4dgs}, radial basis~\cite{stg}, or even learned basis~\cite{dynmf}.
Thanks to the rasterization-based renderings used in 3DGS, most of the aforementioned approaches achieve real-time rendering with promising performance.
However, despite the effort at efficient representation, these methods still require substantial memory and storage.
Among these, we have selected STG~\cite{stg} as our baseline method due to its superior performance, to validate the effectiveness of our universally applicable compact representation.

\section{Compact 3D Gaussian Splatting}
\label{sec:c3dgs}
\subsubsection{Background}
In our approach, we build upon the foundation of 3D Gaussian Splatting (3DGS)~\cite{3dgs}, a point-based representation associated with 3D Gaussian attributes for representing 3D scenes.
$N$ Gaussian are parmeterized by center position $p\in\mathbb{R}^{N\times3}$, opacity $o \in [0,1]^{N}$, 3D scale $s \in \mathbb{R}_{+}^{N\times3}$, 3D rotation represented as a quaternion $r \in\mathbb{R}^{N\times4}$, and spherical harmonics (SH) coefficients $h\in\mathbb{R}^{N\times48}$ (max 3-degrees) for view-dependent color.
The covariance of each Gaussian $\Sigma_n\in\mathbb{R}^{3\times3}$ is positive semi-definite, calculated as follows,
\begin{align}
\label{eq:cov}
    \Sigma_n = {R(r_n)}{S(s_n)}{S(s_n)^T}{R(r_n)^T},
\end{align}
where $n$ is the index of the Gaussian, and $S(\cdot):\mathbb{R}_{+}^{3}\rightarrow \mathbb{R}_{+}^{3\times3}, R(\cdot):\mathbb{R}^{4}\rightarrow  \mathbb{R}^{3\times3}$ stand for diagonal scale matrix from 3D scale and rotation matrix from quaternion, respectively.

To render an image, 3D Gaussians are projected into 2D space by viewing transformation $W$ and Jacobian of the affine approximation of the projective transformation $J$: 
\begin{align}
\label{eq:projcov}
    \Sigma'_n = JW\Sigma_n W^T J^T,
\end{align}
where $\Sigma'_n$ is the projected 2D covariance.
Each pixel color in the image $C(\cdot)$ is then rendered through the alpha composition using colors $c_n$, determined by spherical harmonics under the given view direction, and the final opacity in 2D space $\alpha_n(\cdot)$,
\begin{align}
\label{eq:rast}
    &C(x) = \sum_{k=1}^{\mathcal{N}(x)}c_k\alpha_k(x)\prod_{j=1}^{k-1}(1-\alpha_j(x)),\\
    &\alpha_n(x) = {o_n}\operatorname*{exp}\left(-{1\over2}(x-p'_n)^T{\Sigma'}_n^{-1}(x-p'_n)\right),
\label{eq:projopa}
\end{align}
where $x$ is a coordinate of the pixel to be rendered, $\mathcal{N}(x)$ is the number of Gaussians around $x$, the Gaussians are depth-based sorted given the viewing direction, and $p'_n$ is the projected Gaussian center position, respectively.
3DGS approximates and accelerates this process by introducing the tile-based parallel rasterization pipeline, achieving real-time rendering. For more details, please refer to the original 3DGS paper~\cite{3dgs}.

3DGS constructs initial 3D Gaussians derived from the sparse data points obtained by Structure-from-Motion (SfM), such as COLMAP~\cite{colmap}. 
These Gaussians are cloned, split, pruned, and refined towards enhancing the anisotropic covariance for a precise depiction of the scene.
This training process is based on the gradients from the differentiable rendering without unnecessary computation in empty areas, which accelerates training and rendering.
However, 3DGS's high-quality reconstruction comes at the cost of memory and storage requirements, particularly with numerous Gaussians increased during training and their associated attributes.

\subsubsection{Overall architecture}
Our primary objectives are to 1) reduce the number of Gaussians and 2) represent attributes compactly while retaining the original performance.
To this end, along with the optimization process, we mask out Gaussians that minimally impact performance, as shown in Fig.~\ref{fig:arch}.
For geometry attributes, such as scale and rotation, we propose using a codebook-based method that can fully exploit the limited variations of these attributes.
We represent the color attributes using a grid-based neural field rather than storing their large parameters directly per each Gaussian.
Finally, a small number of Gaussians with compact attributes are then used for the subsequent rendering steps, including projection and rasterization to render images.

\begin{figure}[t]
    \begin{center}
    \includegraphics[width=1.0\linewidth]{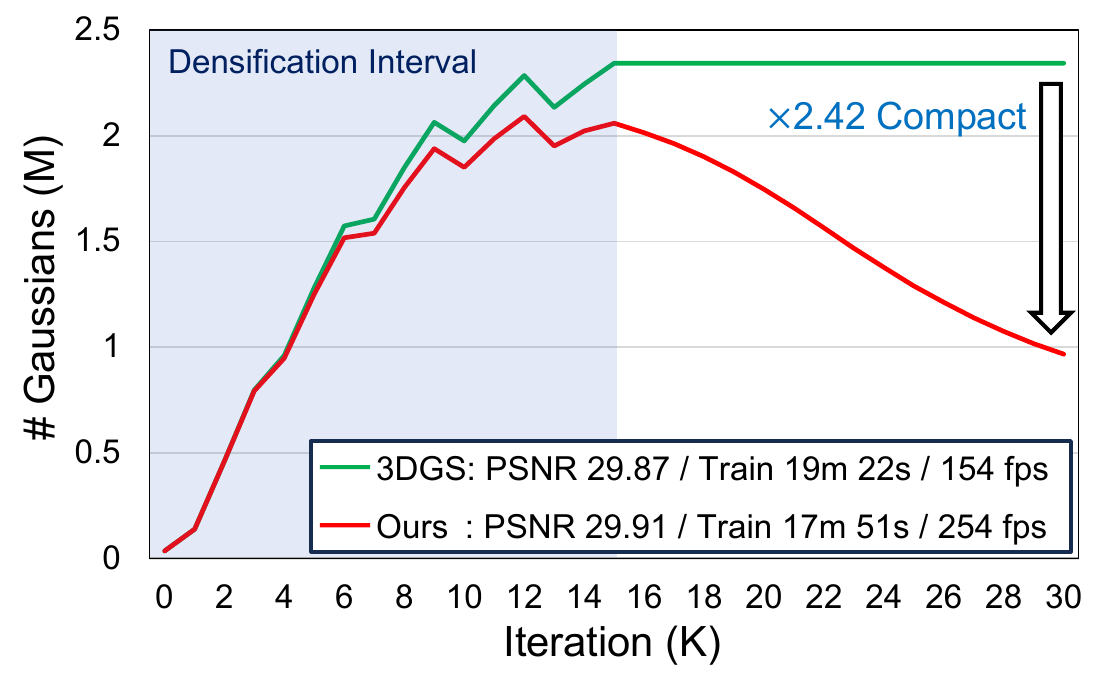}
    \end{center}
    % \vspace{-2em}
    \caption{The number of Gaussians during the training (\textit{Bonsai} scene). `\# Gaussians' denotes the number of Gaussians.}
    % \vspace{-1.5em}
\label{fig:mask}
\end{figure}

\subsection{Gaussian Volume Mask}
3DGS originally densifies Gaussians with large gradients by cloning or splitting.
To regulate the increase in the number of Gaussians, opacities are set to a small number at every specific interval, and after some iterations, those with still minimal opacities are removed.
Although this opacity-based control effectively eliminates some floaters, we empirically found that a significant number of redundant Gaussians still exist ($\times$2.42 Gaussians show similar performance in Fig.~\ref{fig:mask}).
Among them, small-sized Gaussians, due to their minimal volume, have a negligible contribution to the overall rendering quality, often to the point where their effect is essentially imperceptible. 
In such cases, it becomes highly beneficial to identify and remove such unessential Gaussians.

As such, we propose a learnable masking of Gaussians based on their volume as well as transparency.
We apply binary masks not only on the opacities but also on the scale attributes that determine the volume geometry of Gaussians.
We introduce an additional mask parameter $m_n\in \mathbb{R}$, based on which we generate a binary mask $M_n\in \{0,1\}$.
As it is not feasible to calculate gradients from binarized masks, we employ the straight-through estimator~\cite{ste,masked}, formulated as:
\begin{align}
    M_n = \operatorname*{sg}(\mathds{1}[\sigma(m_n) > \epsilon] - \sigma(m_n)) + \sigma(m_n),
\label{eq:bimask}
\end{align}
where $\epsilon$ is the masking threshold, $\operatorname*{sg}(\cdot)$ is the stop gradient operator, and $\mathds{1}[\cdot]$ and $\sigma(\cdot)$ are indicator and sigmoid function, respectively.
By applying the binary mask for the scale and opacity, Eq.~\ref{eq:cov},\ref{eq:projopa} can be reformulated as follows,
\begin{align}
    &\hat{\Sigma}_n = R(r_n)S(M_n s_n)S(M_n s_n)^T R(r_n)^T, \label{eq:scamask}\\
    &\hat{\alpha}_n(x) = M_n{o_n}\operatorname*{exp}\left(-{1\over2}(x-p'_n)^T\hat{\Sigma'}_n^{-1}(x-p'_n)\right),
\label{eq:opamask}
\end{align}
where $\hat{\Sigma'}_n^{-1}$ denotes the projected 2D covariance (Eq. \ref{eq:projcov}) after masking.
This method allows for the incorporation of masking effects based on Gaussian volume and transparency in rendering. Considering both aspects together leads to more effective masking compared to considering either aspect alone.

We balance the accurate rendering and the number of Gaussians eliminated during training by adding masking loss $L_m$ as follows,
\begin{align}
    &L_m = {1\over N} \sum_{n=1}^{N}\sigma({m_n}).
\end{align}

At every densification step, we eliminate Gaussians based on the binary mask.
Furthermore, unlike the original 3DGS, which stops densifying in the middle of the training and retains the number of Gaussians to the end, we consistently mask out throughout the entire training process, reducing unessential Gaussians effectively and ensuring efficient computation with low GPU memory throughout the training phase (Fig.~\ref{fig:mask}).
Once training is completed, we remove the masked Gaussians, so the mask parameter or binary mask does not need to be stored.

\begin{figure*}[t]
    \begin{center}
    \includegraphics[width=1.0\linewidth]{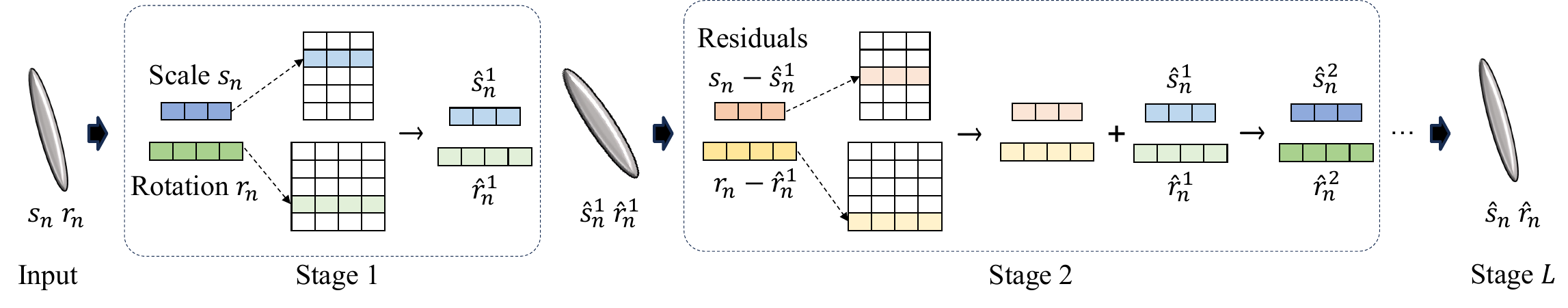}
    \end{center}
    \caption{The detailed process of R-VQ to represent the scale and rotation of Gaussians. In the first stage, the scale and rotation vectors are compared to codes in each codebook, with the closest code identified as the result. In the next stage, the residual between the original vector and the first stage's result is compared with another codebook. This process is repeated up to the final stage, as a result, the selected indices and the codebook from each stage collectively represent the original vector.}
\label{fig:rvq}
\end{figure*}

\subsection{Geometry Codebook}
A number of Gaussians collectively construct a single scene, where similar geometric components can be shared throughout the entire volume.
We have observed that the geometrical shapes of most Gaussians are very similar, showing only minor differences in scale and rotation.
In addition, a scene is composed of many small Gaussians, and each Gaussian primitive is not expected to exhibit a wide range of diversity.
Given this motivation, we propose a codebook learned to represent representative geometric attributes, including scale and rotation, by employing vector quantization (VQ)~\cite{vq}.
As naively applying vector quantization requires a high level of computational complexity and GPU memory~\cite{soundstream}, we adopt residual vector quantization (R-VQ)~\cite{soundstream} that cascades $L$ stages of VQ with codebook size $C$ (Fig.~\ref{fig:rvq}), formulated as follows,
\begin{align}
\label{eq:rvq}
    &\hat{r}_n^{l} = \sum_{j=1}^{l}\mathcal{Z}^j[i_n^j],\,\,\,\, l \in \{1,...,L\}, \\
    &i_n^l = \operatorname*{argmin}_{k} ||\mathcal{Z}^{l}[{k}]-(r_n - \hat{r}_n^{l-1})||_2^2,\,\,\,\, \hat{r}_n^0 = \vec{0}
\end{align}
where $r_n \in\mathbb{R}^{4}$ is the input rotation vector, $\hat{r}^{l}_n \in\mathbb{R}^{4}$ is the output rotation vector after $l$ quantization stages, and $n$ is the index of the Gaussian. $\mathcal{Z}^l\in\mathbb{R}^{C \times 4}$ is the codebook at the stage $l$, $i^l \in \{0,…,C-1\}^{N}$ is the selected indices of the codebook at the stage $l$, and $\mathcal{Z}[i] \in \mathbb{R}^4$ represents the vector at index $i$ of the codebook $\mathcal{Z}$.

The objective function for training the codebooks is as follows,
\begin{align}
    &L_r = {1 \over NC} \sum_{k=1}^{L} \sum_{n=1}^{N} ||\operatorname*{sg}[r_n -\hat{r}_n^{k-1}] - \mathcal{Z}^k[i_n^k]||_2^2,
\end{align}
where $\operatorname*{sg}[\cdot]$ is the stop-gradient operator.
We use the output from the final stage $\hat{r}^{L}$ (we will omit the superscript $L$ for brevity from now onwards), and the R-VQ process is similarly applied to scale $s$ before masking (we also similarly use the objective function for scale $L_s$).

\subsection{Compact View-dependent Color}
Each Gaussian in 3DGS requires 48 of the total 59 parameters to represent SH (max 3 degrees) to model the different colors according to the viewing direction.
Instead of using the naive and parameter-inefficient approach, we propose representing the view-dependent color of each Gaussian by exploiting a grid-based neural field.
To this end, we contract the unbounded positions $p\in\mathbb{R}^{N\times 3}$ to the bounded range, motivated by Mip-NeRF 360~\cite{mip360}, and compute the 3D view direction $d\in\mathbb{R}^{3}$ for each Gaussian based on the camera center point.
We exploit hash grids~\cite{instant-ngp} followed by a tiny MLP to represent color.
Here, we input positions into the hash grids, and then the resulting feature and the view direction are fed into the MLP.
More formally, view-dependent color $c_n(\cdot)$ of Gaussian at position $p_n \in\mathbb{R}^3$ can be expressed as,
\begin{align}
\label{eq:vdcolor}
    &c_n(d) = f(\operatorname*{contract}(p_n), d; \theta),
\end{align}% \vspace{-2em}
\begin{align}
    \operatorname*{contract}(p_n) = \begin{cases} \,\, p_n & ||p_n||\le 1 \\
    \left(2-{1\over||p_n||}\right) \left({p_n\over||p_n||}\right) & ||p_n||>1,
\end{cases}
\end{align}
where $f(\cdot;\theta), \operatorname*{contract}(\cdot):\mathbb{R}^3\rightarrow\mathbb{R}^3$ stand for the neural field with parameters $\theta$, and the contraction function, respectively.
We represent the 0-degree components of SH (the same number of channels as RGB, but not view-dependent) and then convert them into RGB colors due to the slightly increased performance compared to representing the RGB color directly.

\subsection{Training}
\label{sec:train}
Here, we have $N$ Gaussians and their attributes, position $p_n$, opacity $o_n$, rotation $\hat{r}_n$, scale $\hat{s}_n$, and view-dependent color $c_n(\cdot)$, which are used to render images (Eq. ~\ref{eq:rast}). The entire model is trained end-to-end based on the rendering loss $L_{ren}$, the weighted sum of the L1 and SSIM loss between the GT and rendered images.
By adding the loss for masking $L_m$ and geometry codebooks $L_r, L_s$, the overall loss $L$ is as follows,
\begin{align}
    &L = L_{ren} + \lambda_m L_m + L_r + L_s,
\end{align}
where $\lambda_m$ is a hyper-parameter to regularize the number of Gaussians.
To avoid heavy computational costs and ensure fast and optimal training, we initialize the learnable codebooks with K-means and apply R-VQ only during the last 1K training iterations.
Except for that period, we set $L_r,L_s$ to zero.

\begin{figure*}[t]
    \begin{center}
    \includegraphics[width=1.0\linewidth]{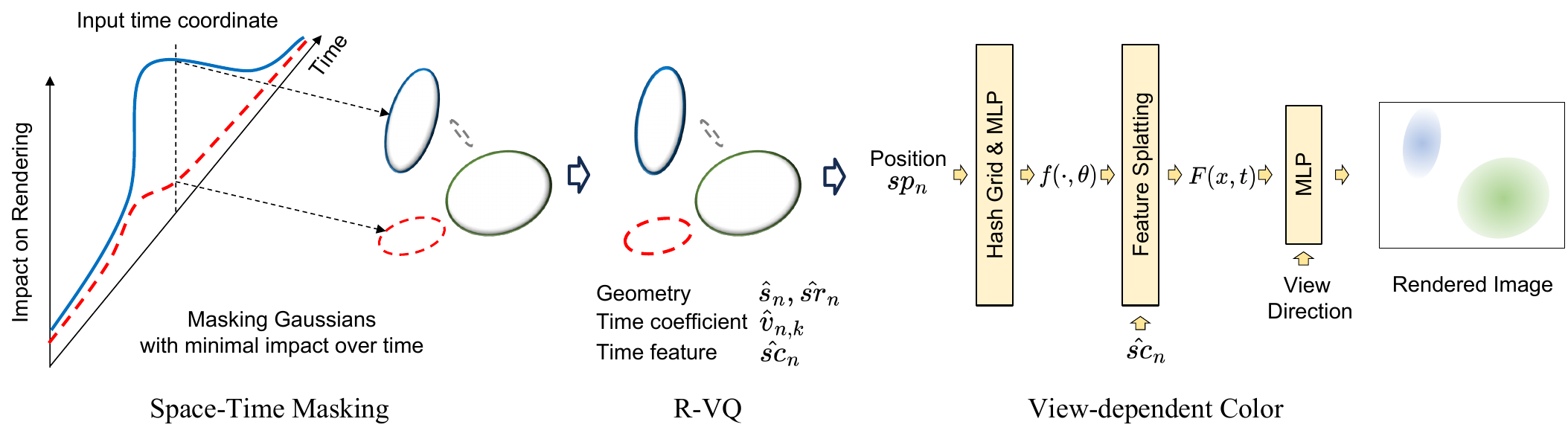}
    \end{center}
    % \vspace{-1.5em}
    \caption{The detailed architecture of our proposed compact Gaussian representation for dynamic scenes.}
    % \vspace{-1.5em}
\label{fig:dyn}
\end{figure*}

\section{Compact 3D Gaussian Splatting for Dynamic Scenes}
Our compact Gaussian representation can be extended to dynamic scenarios. 
We use STG~\cite{stg}, the state-of-the-art method for dynamic scenes, as our baseline model. 
STG is designed to learn space-time Gaussian attributes, including its $no_p$-th and $no_r$-th order polynomial coefficients for moving position $\{u_{n,k}\in\mathbb{R}^3\}_{k=1}^{no_p}$ and rotation $\{v_{n,k}\in\mathbb{R}^4\}_{k=1}^{no_r}$, respectively, and the temporal center and scale $\mu_n, \xi_n\in\mathbb{R}$ as well as the static (time-independent) attributes such as position $sp_n$, rotation $sr_n$, scale $s_n$, opacity $so_n$, and color feature $sc_n$.

STG models the time-conditioned attributes of each Gaussian by introducing a temporal center $\mu_n \in \mathbb{R}$, the time step when each Gaussian is most prominent.
The motions of each Gaussian are represented by learning polynomial coefficients associated with the position $\{u_{n,k}\in\mathbb{R}^3\}_{k=1}^{no_p}$ and rotation $\{v_{n,k}\in\mathbb{R}^4\}_{k=1}^{no_r}$.
At any time $t$, the position $p_n(\cdot)\in\mathbb{R}^3$ and rotation $r_n(\cdot)\in\mathbb{R}^4$ are defined as follows:
\begin{align}
    p_n(t) = sp_n + \sum_{k=1}^{no_p}u_{n,k}(t-\mu_n)^k, \\
    r_n(t) = sr_n + \sum_{k=1}^{no_r}v_{n,k}(t-\mu_n)^k,
\end{align}
where $sp_n\in \mathbb{R}^3$, $sr_n\in \mathbb{R}^4$ are the canonical position and rotation when $t=\mu_n$, and $no_p, no_r$ are the maximum polynomial orders for position and rotation (STG sets $no_p=3, no_r=1$), respectively.
STG remains the scale attribute for each Gaussian $s_n\in \mathbb{R}_+^3$ constant over time.
Therefore, the covariance matrix at time $t$ can be written as,
\begin{align}
\label{eq:dyncov}
&\Sigma_n(t) = R(r_n(t))S(s_n)S(s_n)^T R(r_n(t))^T.
\end{align}

For time-conditioned visibility, STG uses a temporal radial basis function, such that the final projected opacity of each Gaussian $\alpha_n(\cdot,\cdot)$ at the pixel and time coordinates $(x,t)$ is formulated as:
\begin{align}
    &\alpha_n(x, t) = o_n(t)\operatorname*{exp}\left(-{1\over2}(x-p'_n(t))^T\Sigma_n'^{-1}(t)(x-p'_n(t))\right), \\
    &o_n(t) = so_n\operatorname*{exp}\left(-\xi_n|t-\mu_n|^2\right),
\label{eq:dynopa}
\end{align}
where $so_n\in[0,1]$ denotes the time-independent spatial opacity, $\xi_n\in\mathbb{R}$ represents a temporal scale that indicates the effective duration for each Gaussian (i.e., the duration in which its temporal opacity is high), and $p'_n(\cdot), \Sigma_n'(\cdot)$ are the projected center position and covariance at each timestamp.

STG optimizes a 9-dimensional feature for each Gaussian $sc_n \in \mathbb{R}^{9}$ to represent spatial, view-directional, and temporal colors, with 3 dimensions for each, and construct the time-variant color feature of each Gaussian $c_n(t) \in \mathbb{R}^{9}$ as follows,
\begin{align}
    &c_n(t) = \operatorname*{stack}(sc_{n,1:6}, (t-\mu_n)sc_{n,7:9}),
\end{align}
where $sc_{n,1:6}$ is the extracted column vector from the first to $6$-th element of the color feature vector $sc_n$ and $\operatorname*{stack(\cdot,\cdot)}$ operator stacks input vectors into a single vector.

This feature is splatted into the image space through the projection and rasterization process in Eq.~\ref{eq:projcov}-\ref{eq:projopa}, and the splatted feature $F(\cdot,\cdot)$ can be formulated as follows,
\begin{align}
    % &[C_{w,h}^{base},C_{w,h}^{dir},C_{w,h}^{time}]^T = 
    F(x,t) = \sum_{k=1}^{\mathcal{N}(x,t)}c_k(t)\alpha_k(x,t)\prod_{j=1}^{k-1}(1-\alpha_j(x,t)),
\label{eq:dynraster}
\end{align}
where $\mathcal{N}(x,t)$ is the number of Gaussians around $x$ at time $t$, while the Gaussians are depth-based sorted given the viewing direction.
Then the splatted feature $F(x,t)$ is split into $F(x,t)_{1:3}$, $F(x,t)_{4:6}$, and $F(x,t)_{7:9}$, which represent spatial, view-directional, and temporal
color features, respectively, and the final RGB color $C(\cdot,\cdot)$ at pixel and time coordinates $(x,t)$ can be obtained as follows:
\begin{align}
    C(x,t) = F(x,t)_{1:3} + \phi(F(x,t)_{4:6},F(x,t)_{7:9},d),
\label{eq:finalcol}
\end{align}
where $d\in\mathbb{R}^3$ is the viewing direction and $\phi(\cdot, \cdot, \cdot)$ is an MLP for the view- and time-dependent color.
For more details, please refer to the original paper~\cite{stg}.

\subsection{Space-Time Mask}

To eliminate the redundant Gaussians in dynamic scenes, we consider not only the spatial but also the temporal influence of each Gaussian.
We estimate both significances simultaneously by extending the masking strategy (Eq.~\ref{eq:scamask},\ref{eq:opamask}), where the per-Gaussian masks are optimized to reflect their impact on rendering quality over time, as shown in Fig.~\ref{fig:dyn}.
Specifically, we apply the binary mask in Eq.~\ref{eq:bimask} to the time-varying covariance (Eq.~\ref{eq:dyncov}) and opacity (Eq.~\ref{eq:dynopa}), reformulated as follows:
\begin{align}
    &\hat{\Sigma}_n(t) = R(r_n(t))S(M_n s_n)S(M_n s_n)^T R(r_n(t))^T, \\
    &\hat{o}_n(t) = M_n o_n \operatorname*{exp}\left(-\xi_n|t-\mu_n|^2\right).
\end{align}

In the context of static scenes, several methods~\cite{lightgaussian, morton3d} have been suggested to estimate and remove non-essential Gaussians as a post-processing after training, showing promising results. 
However, applying these techniques to dynamic scenes presents greater challenges, as it requires assessing the effectiveness of each Gaussian over the entire time duration. 
Our proposed masking strategy, on the other hand, avoids these complexities thus simplifying the process. It learns the actual rendering impact across all timestamps during training iterations through gradient descent.

\subsection{Compact Attributes}
In Sec.~\ref{sec:c3dgs}, we present the efficient representation for Gaussian attributes using R-VQ and neural fields, depending on the redundancy and continuity of the attributes.
As for static scenes (Eq.~\ref{eq:rvq}), we apply R-VQ to time-invariant geometric attributes (from $s_n, sr_n$ to $\hat{s}_n, \hat{sr}_n$), exploiting the redundancy of them.
In addition, as temporal features exhibit redundancy over time, we also use R-VQ for rotation coefficients $\hat{v}_{n,k}$ and temporal features $\hat{sc}_{n,7:9}$.
However, since the positions require high precision in 3D space and are already compactly represented using polynomial bases, we bypass additional compression for coefficients $u_{n,k}$.

For static color attributes, we similarly exploit the continuity of colors.
We use the neural field $f(\cdot; \theta): \mathbb{R}^{3} \rightarrow \mathbb{R}^{6}$ to represent the spatial and view directional color features $sc_{n,1:6}\in \mathbb{R}^6$ at each canonical position $sp_n$, thus the final color feature $c_n(t)$ can be reformulated as,
\begin{align}
    c_n(t) = \operatorname*{stack}(f(\operatorname*{contract}(sp_n); \ \theta),\ (t-\mu_n)\hat{sc}_{n}),
\end{align}
where $\hat{sc}_{n}\in\mathbb{R}^{3}$ is the R-VQ-applied temporal color feature (we omitted the subscript $_{7:9}$ from $\hat{sc}_{n,7:9}$ for brevity).
Following STG, we splat this feature $c_n(t)$ into the image space, and obtain the final color by using the MLP (Eq.~\ref{eq:finalcol}), as shown in Fig.~\ref{fig:dyn}.

The entire training process resembles that for static scenes described in Sec.~\ref{sec:train}, optimizing rendering loss in conjunction with masking loss and adding further introduced R-VQ losses for temporal coefficients and color.
Also, we use the same strategy for efficient R-VQ applications.

\begin{figure*}[]
    \begin{center}
    \includegraphics[width=1.0\linewidth]{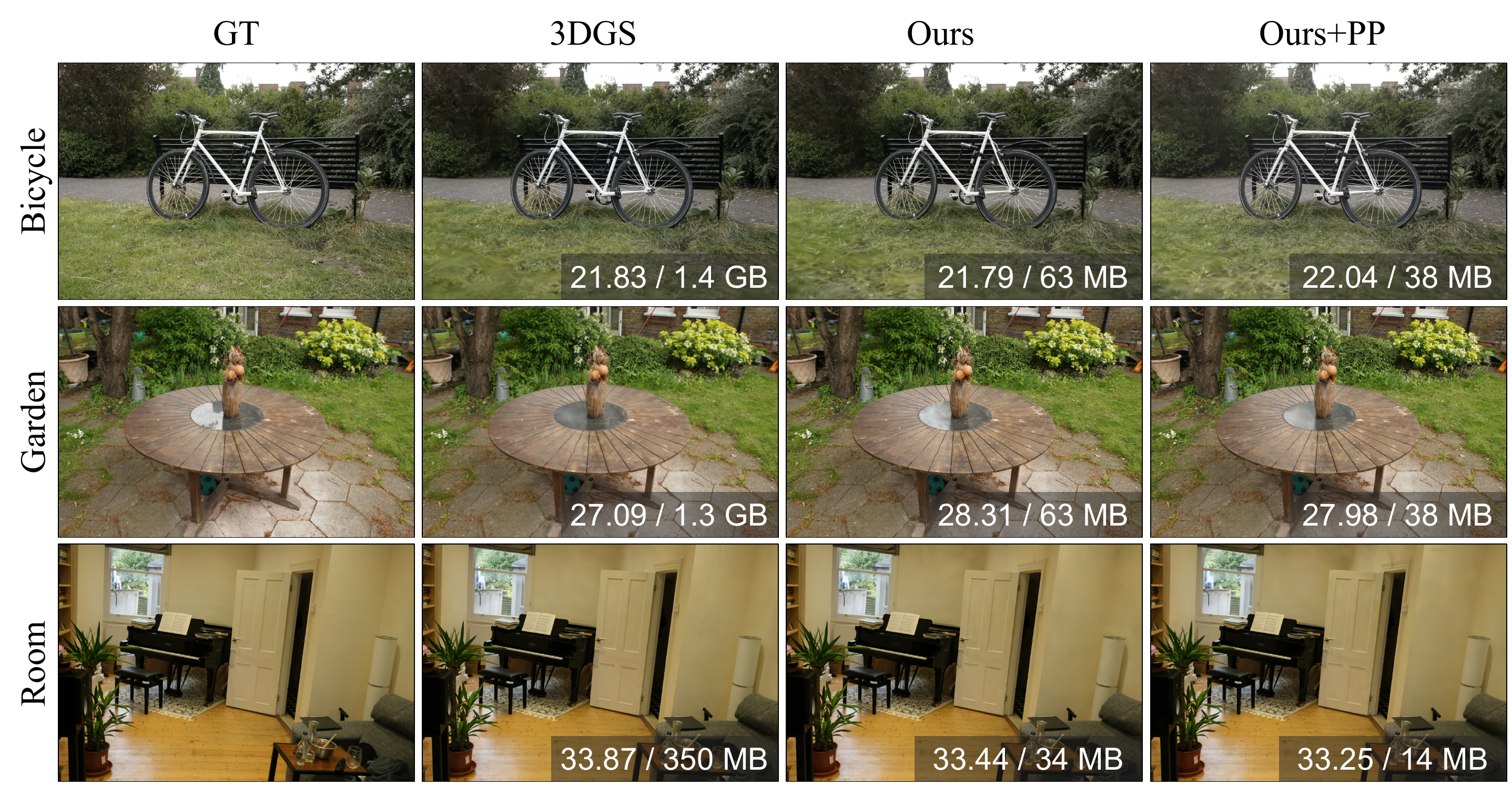}
    \end{center}
    \caption{Qualitative results for static scenes compared to 3DGS. We present the rendering PSNR and storage on the results.}
\label{fig:qual}
\end{figure*}

\begin{table*}[]
\centering
\caption{Quantitative results of the proposed method evaluated on Mip-NeRF 360 and Tanks\&Temples datasets. We reported the numbers of baselines from the original paper (denoted as 3DGS), which were run on an NVIDIA A6000 GPU. For a fair comparison, we re-evaluate 3DGS with the same training configurations as our method using an NVIDIA A100 GPU (denoted as 3DGS*).}
\resizebox{1.0\linewidth}{!}{
\begin{tabular}{lcccccccccccc}
\toprule
Dataset      & \multicolumn{6}{c}{Mip-NeRF 360}                & \multicolumn{6}{c}{Tanks\&Temples}              \\\cmidrule(lr){2-7}\cmidrule(lr){8-13}
Method       & PSNR  & SSIM  & LPIPS & Train  & FPS  & Storage & PSNR  & SSIM  & LPIPS & Train  & FPS  & Storage \\\midrule
Plenoxels    & 23.08 & 0.626 & 0.463 & 25m 49s & 6.79 & 2.1 GB   & 21.08 & 0.719 & 0.379 & 25m 05s  & 13.0 & 2.3 GB   \\
INGP-base    & 25.30 & 0.671 & 0.371 & 05m 37s  & 11.7 & 13 MB    & 21.72 & 0.723 & 0.330 & 05m 26s  & 17.1 & 13 MB    \\
INGP-big     & 25.59 & 0.699 & 0.331 & 07m 30s  & 9.43 & 48 MB    & 21.92 & 0.745 & 0.305 & 06m 59s  & 14.4 & 48 MB    \\
Mip-NeRF 360 & 27.69 & 0.792 & 0.237 & 48h    & 0.06 & 8.6 MB   & 22.22 & 0.759 & 0.257 & 48h    & 0.14 & 8.6 MB   \\
3DGS         & 27.21 & 0.815 & 0.214 & 41m 33s & 134  & 734 MB   & 23.14 & 0.841 & 0.183 & 26m 54s & 154  & 411 MB   \\\midrule
3DGS*         & 27.46 & 0.812 & 0.222 & 24m 07s & 120  & 746 MB   & 23.71 & 0.845 & 0.178 & 13m 51s & 160  & 432 MB   \\
Ours         & 27.08 & 0.798 & 0.247 & 33m 06s & \textbf{128}  & \textbf{48.8 MB}  & 23.32 & 0.831 & 0.201 & 18m 20s & \textbf{185}  & \textbf{39.4 MB} \\
Ours+PP         & 27.03 & 0.797 & 0.247 & - & -  & \textbf{26.2 MB}  & 23.32 & 0.831 & 0.202 & - & -  & \textbf{18.9 MB}
\\\bottomrule
\end{tabular}}
\label{tab:qual1}
\end{table*}

\section{Experiment}

\subsection{Implementation Details}
\subsubsection{Static scenes} We tested our approach on three real-world datasets (Mip-NeRF 360~\cite{mip360}, Tanks\&Temples~\cite{tnt}, and Deep Blending~\cite{db}) and a synthetic dataset (NeRF-Synthetic~\cite{nerf}).
Following 3DGS, we chose two scenes from Tanks\&Temples and Deep Blending.
We retained all hyper-parameters of 3DGS and trained models for 30K iterations, and we set the codebook size $C$ and the number of stages $L$ of R-VQ for geometry to 64 and 6, respectively.
The neural field for view-dependent color uses hash grids with 2-channel features across 16 different resolutions (16 to 4096) and a following 2-layer 64-channel MLP.
Due to the different characteristics between the real and synthetic scenes, we adjusted the maximum hash map size and the hyper-parameters for learning the neural field and the mask.
For the real scenes, we set the max size of hash maps to $2^{19}$, the control factor for the number of Gaussians $\lambda_m$ to $5e^{-4}$, and the learning rate of the mask parameter and the neural fields to $1e^{-2}$.
The learning rate of the neural fields is decreased at 5K, 15K, and 25K iterations by multiplying a factor of $0.33$.
For the synthetic scenes, the maximum hash map size and the control factor $\lambda_m$ were set to $2^{16}$ and $4e^{-3}$, respectively.
The learning rate of the mask parameter and the neural fields were set to $1e^{-3}$, where the learning rate of the neural fields was reduced at 25K iterations with a factor of $0.33$.
\subsubsection{Dynamic scenes}
We trained models for 25K iterations using two real-world multi-view video datasets (DyNeRF~\cite{Li_2022_CVPR} and Technicolor~\cite{technicolor}), retaining all other hyper-parameters of STG.
We set the codebook size $C$ to 256 and the number of stages $L$ (geometry, temporal attributes) to (4,3) and (5,4) for DyNeRF and Technicolor datasets, respectively.
The max hash map sizes were set to $2^{14}$, $2^{16}$ for DyNeRF and Technicolor datasets, respectively, and the learning rate of the neural fields is decreased at 3, 6, 9, 12, 18, and 21K iterations by multiplying a factor of $0.33$.
The other settings for the neural field structure and learning rates remained the same as those we used for real-world static scenes.

\begin{table}[t]
\caption{Quantitative results of the proposed method evaluated on Deep Blending dataset. We re-evaluate 3DGS* under the same configurations with our method.}
\resizebox{1.0\linewidth}{!}{
\begin{tabular}{lcccccc}
\toprule
Dataset      & \multicolumn{6}{c}{Deep Blending}                \\\cmidrule(lr){2-7}
Method       & PSNR  & SSIM  & LPIPS & Train   & FPS  & Storage \\\midrule
Plenoxels    & 23.06 & 0.795 & 0.510 & 27m 49s & 11.2 & 2.7 GB  \\
INGP-base    & 23.62 & 0.797 & 0.423 & 06m 31s & 3.26 & 13 MB   \\
INGP-big     & 24.96 & 0.817 & 0.390 & 08m 00s & 2.79 & 48 MB   \\
Mip-NeRF 360 & 29.40 & 0.901 & 0.245 & 48h     & 0.09 & 8.6 MB  \\
3DGS         & 29.41 & 0.903 & 0.243 & 36m 02s & 137  & 676 MB  \\\midrule
3DGS*         & 29.46 & 0.900 & 0.247 & 21m 52s & 132  & 663 MB  \\
Ours         & \textbf{29.79} & \textbf{0.901} & 0.258 & 27m 33s & \textbf{181}  & \textbf{43.2 MB}\\
Ours+PP         & 29.73 & 0.900 & 0.258 & - & -  & \textbf{21.6 MB}\\\bottomrule
\end{tabular}}
\label{tab:qual2}
\end{table}

\begin{table}[t]
\centering
\caption{Quantitative results of the proposed method evaluated on NeRF-Synthetic dataset. * denotes the reported value in the original paper.}
\resizebox{1.0\linewidth}{!}{
\begin{tabular}{lcccc}\toprule
Dataset & \multicolumn{4}{c}{NeRF-Synthetic}                     \\
\cmidrule(lr){2-5}
Method  & PSNR  & Storage & Train & FPS \\\midrule
3DGS    & 33.32* &  68.1 MB                        & 6m 14s & 359 \\
Ours    & 33.33 & \textbf{5.55 MB} ($\times$0.08)                         & 8m 04s ($\times$1.29)  & \textbf{545} ($\times$1.52)\\
Ours+PP    & 32.88 & \textbf{2.47 MB} ($\times$0.04)                         & -  & - \\\bottomrule
\end{tabular}}
\label{tab:qual3}
\end{table}

\subsubsection{Post-processing}
For the end-to-end trained models with the proposed method (denoted as Ours), we stored the position (including coefficients for position) and scalar attributes (opacity of 3DGS; opacity, temporal center, and temporal scale of STG) with 16-bit precision using half-tensors.
Additionally, we implemented straightforward post-processing techniques on the model attributes, a variant we denote as Ours+PP. These post-processing steps include:
\begin{itemize}
    \item Applying 8-bit min-max quantization to hash grid parameters and scalar attributes.
    \item Pruning hash grid parameters with values below 0.1.
    \item Sorting Gaussians in Morton order~\cite{morton3d}.
    \item Applying Huffman encoding~\cite{huffman} on the 8-bit quantized values (hash parameters and scalar attributes) and R-VQ indices, and compressing the results using DEFLATE~\cite{deflate}.
\end{itemize}

\begin{figure*}[]
    \begin{center}
    \includegraphics[width=0.9\linewidth]{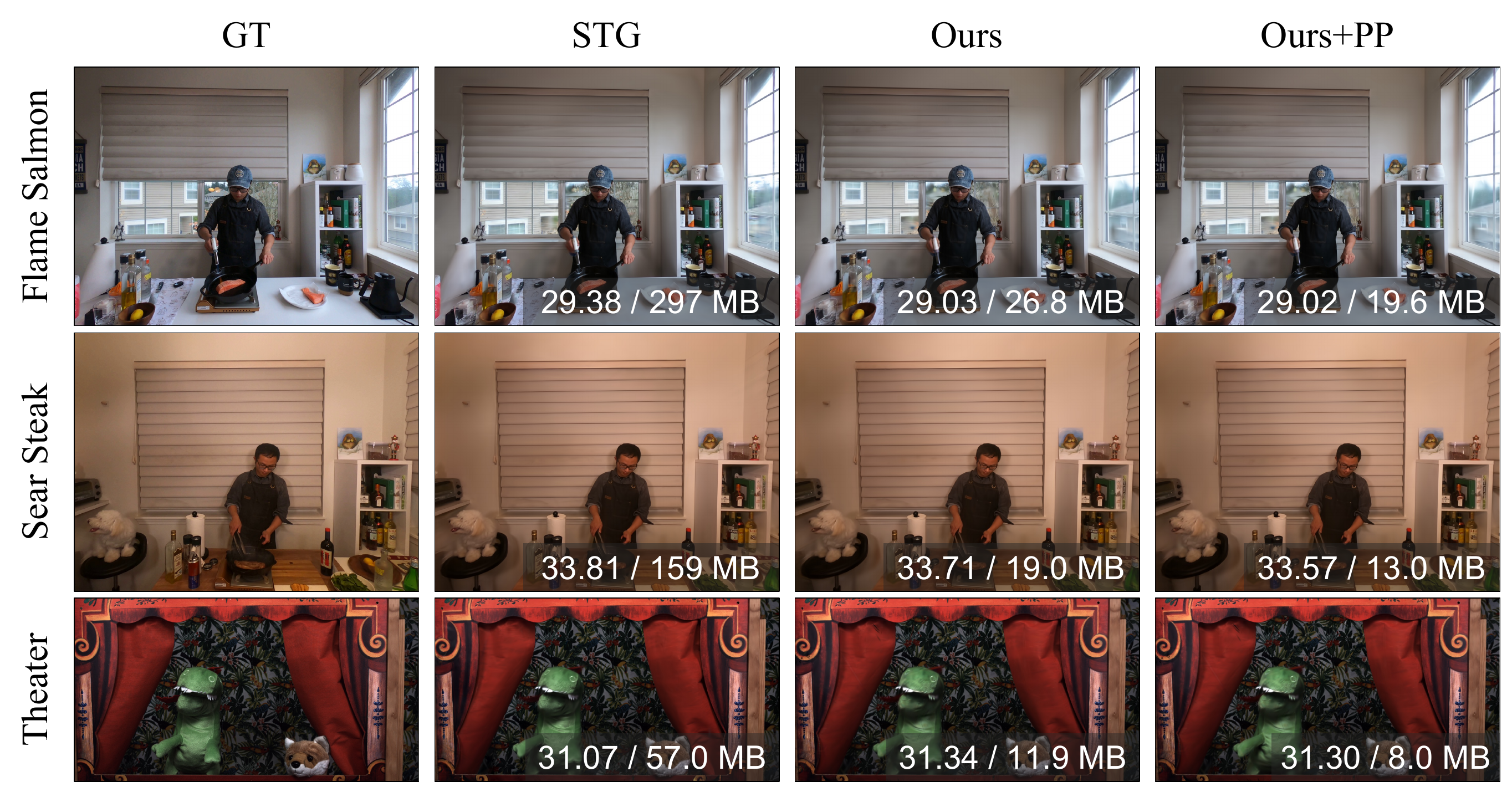}
    \end{center}
    % \vspace{-2em}
    \caption{Qualitative results for dynamic scenes compared to STG. We present the rendering PSNR and storage on the results.}
    % % \vspace{-0.5em}
\label{fig:qual_4d}
\end{figure*}

\subsection{Static Scene Representation}
\subsubsection{Real-world scenes} \Cref{tab:qual1} and \Cref{tab:qual2} show the qualitative results evaluated on real-world scenes.
Across datasets, our approach achieves high reconstruction performance comparable to 3DGS while drastically reducing the overall size and accelerating rendering.
Especially for the Deep Blending dataset (\Cref{tab:qual2}), our method even outperforms the original 3DGS in terms of visual quality (measured in PSNR and SSIM), achieving state-of-the-art performance with the fastest rendering speed as well as compactness (almost 40\% faster rendering and over 15$\times$ compactness compared to 3DGS).
Comparing our method to 3DGS, the qualitative results in Fig.~\ref{fig:qual} further demonstrate our method's high-quality rendering with substantially less size.

\subsubsection{Synthetic scenes} We also evaluate our method on synthetic scenes.
As 3DGS has proven its effectiveness in improving visual quality, rendering speed, and training time compared to other baselines, we focus on comparing our method with 3DGS, highlighting the improvements from our method. As shown in \Cref{tab:qual3}, although our approach requires slightly more training duration, we achieve over 10$\times$ compression and 50\% faster rendering compared to 3DGS, maintaining high-quality reconstruction.

\subsubsection{Post-processings} With post-processings, our model can be further downsized by over 40 \% regardless of the dataset. 
Consequently, we achieve more than 28$\times$ compression from 3DGS (Mip-NeRF 360), while maintaining high performance.

\begin{table}[t]
\caption{Quantitative results of the proposed method evaluated on DyNeRF dataset. We re-evaluate STG* under the same configurations with our method.}
\label{tab:4d}
\resizebox{\linewidth}{!}{
\begin{tabular}{lcccccc}
\toprule
Dataset      & \multicolumn{6}{c}{DyNeRF}                                                 \\\cmidrule(lr){2-7}
Method       & PSNR           & SSIM           & SSIM' & LPIPS          & FPS  & Storage          \\\midrule
NeRFPlayer   & 30.69          & 0.932          & -     & 0.111          & 0.05 & 5.1 GB           \\
K-Planes     & 31.63          & -              & 0.964 & -              & 0.3  & 311 MB           \\
MixVoxels-L  & 31.34          & -              & 0.966 & 0.096          & 37.7 & 500 MB           \\
Dynamic 3DGS & 30.67          & 0.930          & 0.962 & 0.099          & 460  & 2.7 GB           \\
4DGS         & 31.15          & -              & 0.968 & 0.049          & 30   & 90 MB            \\
STG-Lite     & 31.59          & 0.944          & 0.968 & 0.047          & 310  & 103 MB           \\
STG          & \textbf{32.05} & 0.946          & 0.970 & \textbf{0.044} & 140  & 200 MB           \\\midrule
STG*         & 31.94          & \textbf{0.948} & 0.971 & 0.046          & 181  & 197 MB           \\
Ours         & 31.73          & 0.945          & 0.969 & 0.053          & 186  & 21.8 MB          \\
Ours+PP      & 31.69          & 0.945          & 0.969 & 0.054          & -    & \textbf{15.4 MB} \\\bottomrule
\end{tabular}}
\end{table}

\begin{table}[t]
\centering
\caption{Quantitative results of the proposed method evaluated on Technicolor dataset. We re-evaluate STG* under the same configurations with our method.}
\label{tab:techni}
\begin{tabular}{lccccc}
\toprule
Dataset      & \multicolumn{5}{c}{Technicolor}                                                 \\\cmidrule(lr){2-6}
Method       & PSNR          & SSIM           & LPIPS          & FPS          & Storage/Fr       \\\midrule
DyNeRF    & 31.8          & -              & 0.140          & 0.02         & 0.6 MB           \\
HyperReel & 32.7          & 0.906          & 0.109          & 4.0          & 1.2 MB           \\
STG       & \textbf{33.6} & \textbf{0.920} & 0.084          & 86.7         & 1.1 MB           \\\midrule
STG*      & 33.5          & \textbf{0.920} & \textbf{0.083} & 105          & 1.3 MB           \\
Ours+PP   & 33.1         & 0.910          & 0.098          & \textbf{116} & \textbf{0.16 MB} \\\bottomrule
\end{tabular}
\end{table}

\subsection{Dynamic Scene Representation}
\Cref{tab:4d} and \Cref{tab:techni} show the qualitative results evaluated on DyNeRF and Technicolor datasets.
For both datasets, our approach achieves high-performance representation comparable to STG while significantly reducing the storage. 
Especially for the DyNeRF dataset, our method achieves over 9$\times$ compactness compared to STG, even though STG has already been designed for compact representation.
With post-processings, our model can be further downsized by almost 30 \%, consequently, we achieve more than 12$\times$ compression from STG, while maintaining high performance.
Fig.~\ref{fig:qual_4d} illustrates the qualitative results compared to STG, highlighting our method's high-quality reconstruction with significantly reduced size.

\begin{table*}[]
\centering
\caption{Ablation study on the proposed contributions, masking, color representation, geometry codebook, and half tensor for positions and opacities. `\#Gauss' means the number of Gaussians.}
\begin{tabular}{ccccccccccccccc}\toprule
\multicolumn{5}{c}{Method \textbackslash Dataset} & \multicolumn{5}{c}{Playroom}                 & \multicolumn{5}{c}{Bonsai}                   \\
\cmidrule(lr){1-5}\cmidrule(lr){6-10}\cmidrule(lr){11-15}
Mask    & Col    & Geo  & Half   & Post  & PSNR  & Train time & \#Gauss & Storage & FPS & PSNR  & Train time & \#Gauss & Storage & FPS \\\midrule
\multicolumn{5}{c}{3DGS}                          & 29.87 & 19m 22s    & 2.34 M  & 553 MB  & 154 & 32.16 & 19m 18s    & 1.25 M  & 295 MB  & 200 \\
\checkmark          &             &            &      &     & 29.91 & 17m 51s    & 967 K   & 228 MB  & 254 & 32.22 & 18m 50s    & 643 K   & 152 MB  & 247 \\
\checkmark          & \checkmark          &            &     &      & 30.33 & 23m 56s    & 770 K   & 59 MB   & 210 & 32.08 & 23m 09s    & 592 K   & 51 MB   & 196 \\
\checkmark          & \checkmark          & \checkmark         &     &      & 30.33 & 24m 58s    & 761 K   & 44 MB   & 204 & 32.08 & 24m 06s    & 598 K   & 40 MB   & 198 \\
\checkmark        & \checkmark         & \checkmark        & \checkmark   &    & 30.32 & 24m 35s    & 778 K   & 38 MB   & 206 & 32.08 & 24m 16s    & 601 K   & 35 MB   & 196 \\
\checkmark        & \checkmark         & \checkmark        & \checkmark   &  \checkmark  & 30.30 & -    & -   & 17 MB   & - & 31.98 & -    & -   & 15 MB   & - \\\bottomrule
\end{tabular}
\label{tab:abl}
\end{table*}

\begin{table*}[]
\centering
\caption{Ablation study on the proposed contributions, masking, color representation, temporal codebook, geometry codebook, and half tensor for positions and scalar attributes, for dynamic scenes. `\#Gauss' means the number of Gaussians.}
\begin{tabular}{cccccccccccccccc}
\toprule
\multicolumn{6}{c}{Method \textbackslash Dataset}                     & \multicolumn{5}{c}{Painter}             & \multicolumn{5}{c}{Cut Roasted Beef}    \\\cmidrule(lr){1-6}\cmidrule(lr){7-11}\cmidrule(lr){12-16}
Mask      & Color     & Time      & Geo       & Half      & Post      & PSNR  & SSIM  & \#Gauss & Storage & FPS & PSNR  & SSIM  & \#Gauss & Storage & FPS \\\midrule
\multicolumn{6}{c}{STG}                                               & 36.21 & 0.929 & 553 K   & 84.1 MB & 110 & 33.43 & 0.959 & 1.00 M& 152 MB  & 181 \\
\checkmark &           &           &           &           &           & 36.29 & 0.927 & 145 K   & 22.0 MB & 132 & 33.32 & 0.958 & 342 K   & 51.9 MB & 220 \\
\checkmark & \checkmark &           &           &           &           & 36.45 & 0.925 & 121 K   & 19.2 MB & 127 & 33.11 & 0.956 & 287 K   & 42.7 MB & 208 \\
\checkmark & \checkmark & \checkmark &           &           &           & 36.28 & 0.923 & 132 K   & 16.4 MB & 122 & 33.06 & 0.955 & 286 K   & 33.0 MB & 210 \\
\checkmark & \checkmark & \checkmark & \checkmark &           &           & 36.22 & 0.923 & 132 K   & 14.0 MB & 124 & 33.05 & 0.955 & 286 K   & 28.6 MB & 212 \\
\checkmark & \checkmark & \checkmark & \checkmark & \checkmark &           & 36.35 & 0.923 & 132 K   & 10.2 MB & 115 & 33.09 & 0.955 & 286 K   & 19.4 MB & 208 \\
\checkmark & \checkmark & \checkmark & \checkmark & \checkmark & \checkmark & 36.35 & 0.923 & -       & 6.56 MB & -   & 33.03 & 0.955 & -       & 13.3 MB & -  \\\bottomrule
\end{tabular}
\label{tab:abl4d}
\end{table*}

\begin{figure*}[]
    \begin{center}
    \includegraphics[width=1.0\linewidth]{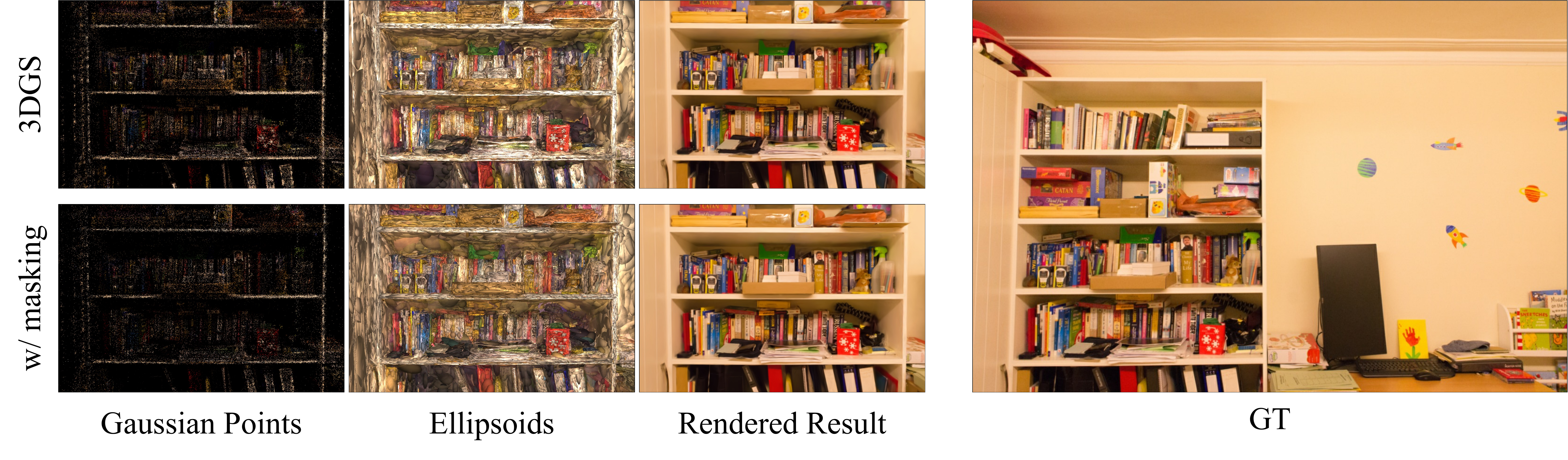}
    \end{center}
    \caption{Effect of the proposed learnable volume masking, compared to the original 3DGS. We visualize Gaussian center points, ellipsoids, and rendered results using the \textit{Playroom} scene.}
\label{fig:vis_mask}
\end{figure*}

\begin{figure*}[]
    \begin{center}
    \includegraphics[width=0.91\linewidth]{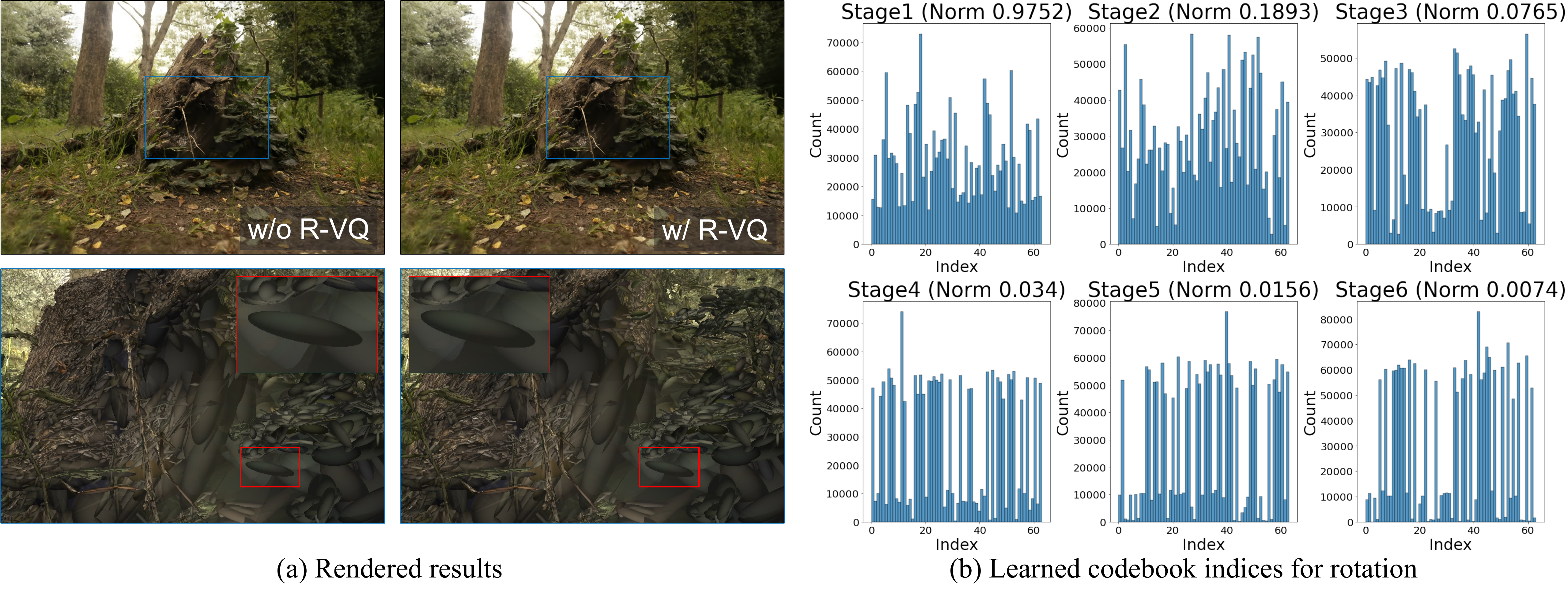}
    \end{center}
    \caption{Effect of the proposed geometry codebook. We visualize (a) ellipsoids and rendered results, and (b) learned codebook indices for rotation using the \textit{Stump} scene. `Norm' denotes the average norm of all code vectors in each codebook (representing magnitude).}
\label{fig:vis_rvq}
\end{figure*}

\subsection{Ablation Study}

\subsubsection{Learnable masking}
As shown in \Cref{tab:abl}, the proposed volume-based masking significantly reduces the number of Gaussians while retaining (even slightly increasing) the visual quality, demonstrating its effectiveness in removing redundant Gaussians.
The reduced Gaussians show several additional advantages: reducing training time, storage, and testing time.
Specifically on the \textit{Playroom} scene, our proposed masking method shows a 140\% increase in storage efficiency and a 65\% increase in rendering speed.
Furthermore, our method can remove redundant Gaussians effectively regarding space-time redundancy in dynamic scenarios (\Cref{tab:abl4d}). 
Especially for the \textit{Painter} scene, we reduce almost 75\% of Gaussians and increase rendering speed by 20\% without sacrificing rendering quality.

Fig.~\ref{fig:vis_mask} further illustrates the effect of the masking method on Gaussians and the resulting images.
Despite the noticeable reduction in the number of Gaussians, as evidenced by the sparser points in the visualization, the quality of the rendered results remains high, with no visible differences. 
These results demonstrate the effectiveness and efficiency of the proposed method, both quantitatively and qualitatively.

\subsubsection{Compact color representation}
For static scenes, the proposed color representation based on the neural field offers more than a threefold improvement in storage efficiency with a slightly reduced number of Gaussians compared to directly storing high-degree SH, despite necessitating slightly more time for training and rendering. 
Nonetheless, when compared to 3DGS, the proposed color representation with the masking strategy demonstrates either a faster or comparable rendering speed.

When applying our compact color representation to STG, we achieve an additional 14\% compression from the model with masked Gaussians. This is significant, considering that STG is already focused on compactness for color representation.

\subsubsection{Codebook approach}
Our proposed geometry codebook approach achieves a reduction in storage requirements by approximately 30\% while maintaining the reconstruction quality, training time, and rendering speed, as shown in \Cref{tab:abl}. Furthermore, we have validated that this codebook approach effectively reduces the storage requirements for temporal attributes in dynamic scenes as well as geometry (\Cref{tab:abl4d}).

To analyze the actual effectiveness of the codebook-based approach, we showcase the geometry of the Gaussians, as shown in Fig.~\ref{fig:vis_rvq}-(a).
We can observe that the majority of Gaussians maintain their scales and rotations regardless of R-VQ, with only minor differences in a few that are hardly noticeable.
We also explore the patterns of learned indices across each stage of R-VQ, shown in Fig.~\ref{fig:vis_rvq}-(b).
The lower stages exhibit relatively even distributions with large magnitudes of codes.
As the stages progress, the distribution gets uneven and the magnitude of codes decreases, indicating the residuals of each stage have been reduced and trained to represent geometry.

\begin{table}[t]
\caption{Avarge storage (MB) for each Gaussian attribute, evaluated on Mip-NeRF 360 dataset. f, 8b, H, and P mean floating-point, min-max quantization to 8-bit, Huffman encoding, and pruning parameters below 0.1, respectively. The value in parentheses indicates the result after DEFLATE compression.}
\resizebox{1.0\linewidth}{!}{
\begin{tabular}{lccccccc}
    \toprule
     & Pos. & Opa. & Sca. & Rot. & \multicolumn{2}{c}{Col.} & Tot.  \\\midrule
    \multirow{2}{*}{3DGS} & \multicolumn{6}{c}{32f} & \multirow{2}{*}{746}  \\\cmidrule(lr){2-7}
     & 37.9  & 12.6  & 37.9  & 50.6  & \multicolumn{2}{c}{606.9} &   \\\midrule
    \multirow{2}{*}{Ours} & \multicolumn{2}{c}{16f} & \multicolumn{2}{c}{R-VQ} & Hash(16f) & MLP(16f) & \multirow{2}{*}{48.8}  \\\cmidrule(lr){2-3}\cmidrule(lr){4-5}\cmidrule(lr){6-6}\cmidrule(lr){7-7}
     & 8.3  & 2.8  & 6.3  & 6.3  & 25.2 & 0.016  &   \\\midrule
    \multirow{2}{*}{\begin{tabular}[l]{@{}l@{}}Ours\\+PP\end{tabular}} & 16f & 8b+H & \multicolumn{2}{c}{+H} & +8b+P+H & MLP(16f) & \multirow{2}{*}{\begin{tabular}[c]{@{}c@{}}29.1\\\textit{(26.2)}\end{tabular}} \\\cmidrule(lr){2-2}\cmidrule(lr){3-3}\cmidrule(lr){4-5}\cmidrule(lr){6-6}\cmidrule(lr){7-7}
     & 8.3  & 1.2  & 5.9  & 6.2  & 7.4  & 0.016  & 
    \\\bottomrule
    \end{tabular}}
\label{tab:size}
\end{table}

\subsubsection{Effect of post-processing }
As shown in \Cref{tab:abl},\ref{tab:abl4d}, our post-processing techniques significantly reduce the storage requirement without performance drop, both for static and dynamic scenes.
Additionally, \Cref{tab:size} describes the size of each attribute with and without the application of post-processing techniques for static scenes. 
While our end-to-end trainable framework demonstrates significant effectiveness, it requires relatively large storage for color representation. 
Nonetheless, as indicated in the table, this can be effectively reduced through simple post-processing.

\section{Conclusion}
We have proposed a compact 3D Gaussian representation for both static and dynamic 3D scenes, reducing the number of Gaussians without sacrificing visual quality through a novel learnable masking.
Furthermore, this work proposed combining the neural field and exploiting the learnable codebooks to represent Gaussian attributes compactly.
Through extensive experiments, our approach demonstrated more than 25$\times$ and 12$\times$ reduction in storage compared to 3DGS and STG, respectively, and an increase in rendering speed while retaining high-quality reconstruction.
These results set a new benchmark with high visual quality, compactness, fast training, and real-time rendering for both static and dynamic radiance fields. 
Our framework thus stands as a comprehensive solution, paving the way for broader adoption and application in various fields requiring efficient and high-quality 3D scene representation.

\bibliographystyle{IEEEtran}
\bibliography{main}

\begin{IEEEbiography}[{\includegraphics[trim={0cm 0cm 0cm 0.2cm},width=1in,height=1.25in,clip, keepaspectratio]{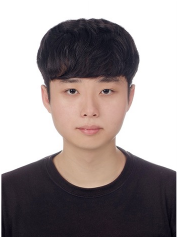}}]{Joo Chan Lee}
(Graduate Student Member, IEEE) received the B.S. degree in information and communication engineering from Inha University in 2020. He is currently pursuing the Ph.D. degree in artificial intelligence from Sungkyunkwan University.
His current research interests include the areas of computer vision, graphics, and machine learning.
\end{IEEEbiography}

\begin{IEEEbiography}[{\includegraphics[trim={0.45cm 0cm 0.45cm 0cm},width=1in,height=1.25in,clip,keepaspectratio]{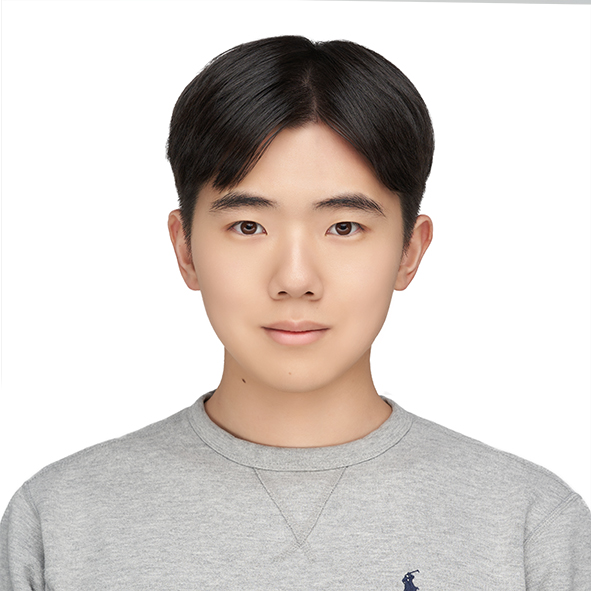}}]{Daniel Rho}
    is currently a Ph.D. student at the University of North Carolina at Chapel Hill. He received an M.S. degree in Artificial Intelligence from Sungkyunkwan University, South Korea, in 2022, and a double major in Economics and Computer Science from the same university in 2020. His research interests include neural rendering, computer graphics, and machine learning.
\end{IEEEbiography}

\begin{IEEEbiography}[{\includegraphics[width=1in,height=1.25in,clip,keepaspectratio]{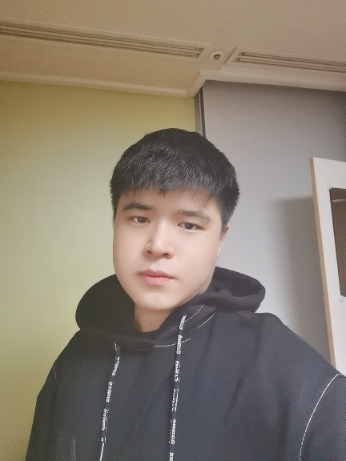}}]{Xiangyu Sun}
received the bachelor’s and master’s degrees from Huazhong University of Science and Technology, China, in 2018 and 2021. He is currently working toward the Ph.D. degree with the VSC Lab, Sungkyunkwan University. His research interests include 3D reconstruction, neural radiance field and 3D generative model.\end{IEEEbiography}

\begin{IEEEbiography}[{\includegraphics[trim={0.5cm 0cm 0.5cm 0.3cm}, width=1in,height=1.25in,clip]{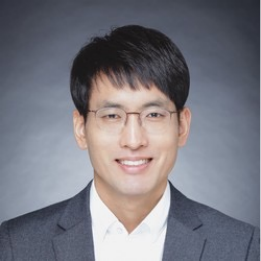}}]{Jong Hwan Ko}
(Member, IEEE) received the dual B.S. degrees in computer science and engineering and mechanical and aerospace engineering and the M.S. degree in electrical engineering and computer science from Seoul National University, and the Ph.D. degree from the School of Electrical and Computer Engineering, Georgia Tech, in 2018. During his seven years of research experience at the Agency for Defense Development (ADD) in South Korea, he conducted advanced research on the design and performance analysis of military wireless sensor networks. He joined Sungkyunkwan University (SKKU), South Korea, as an Assistant Professor. His research interests include design of low-power image sensor systems and deep-learning accelerators for efficient image/audio processing. He has received the Best Paper Award from the International Symposium on Low Power Electronics and Design (ISLPED), in 2016.
\end{IEEEbiography}

\begin{IEEEbiography}[{\includegraphics[trim={1.2cm 0cm 1.2cm 0cm},width=1in,height=1.25in,clip,keepaspectratio]{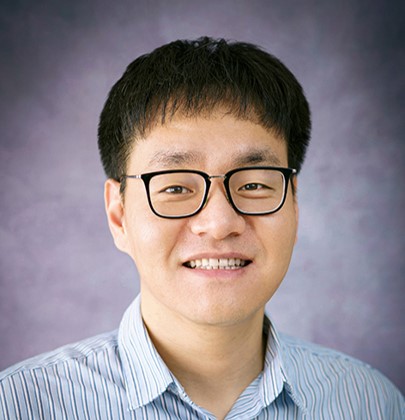}}]{Eunbyung Park}
(Member, IEEE) received a B.S. degree in computer science from Kyung Hee University in 2009, an M.S. degree in computer science from Seoul National University in 2011, and a Ph.D. degree in computer science from the University of North Carolina at Chapel Hill in 2019. He is currently an assistant professor in the Department of Electronic and Electrical Engineering at Sungkyunkwan University (SKKU), South Korea. Before joining SKKU, he was a research scientist at Nuro and an applied scientist at Microsoft. During his doctoral study, he has worked at various research institutes, including Google DeepMind, Microsoft Research, Adobe Research, and HP Labs. His current research interests include computer vision and machine learning and its applications to visual and scientific computing.\end{IEEEbiography}

\end{document}